\documentclass[12pt,a4paper]{article}

\usepackage{essv}
\usepackage[T1]{fontenc}
\usepackage[utf8]{inputenc}
\usepackage{babel}
\usepackage[font=small,labelfont=bf]{caption}
\usepackage{subcaption,siunitx,booktabs}
\usepackage{floatrow}
\usepackage{float}
\usepackage[acronym]{glossaries}
\makeglossaries
\newacronym[plural=DNNs,firstplural=deep neural networks (DNNs)]{DNNs}{DNNs}{deep neural networks}
\newacronym[plural=CNNs,firstplural=convolutional neural networks (CNNs)]{CNNs}{CNNs}{convolutional neural networks}
\newacronym[plural=RNNs,firstplural=recurrent neural networks (RNNs)]{RNNs}{RNNs}{recurrent neural networks}
\newacronym[plural=TDNNs,firstplural=time delay neural networks (TDNNs)]{TDNNs}{TDNNs}{time delay neural networks}
\newacronym[plural=DenseNets,firstplural=densely connected convolutional networks (DenseNets)]{DenseNets}{DenseNets}{densely connected convolutional networks}
\newacronym[plural=DenseNets-AD,firstplural=densely connected convolutional networks with domain adversarial learning (DenseNets-AD)]{DenseNets-Dal}{DenseNets-Dal}{densely connected convolutional networks with domain adversarial learning}
\newacronym[plural=LACE,firstplural=Deep Convolutional Neural Networks with Layer-wise Context Expansion and Attention architecture (LACE)]{LACE}{LACE}{Deep Convolutional Neural Networks with Layer-wise Context Expansion and Attention architecture}
\newacronym[plural=LSTMs,firstplural=Long Short-Term Memory Networks (LSTMs)]{LSTMs}{LSTMs}{Long Short-Term Memory Networks}
\newacronym[plural=ASR,firstplural=Automatic Speech Recognition (ASR)]{ASR}{ASR}{Automatic Speech Recognition}
\newacronym[plural=SWB,firstplural=Switchboard Task (SWB)]{SWB}{SWB}{Switchboard Task}
\newacronym[plural=RM,firstplural=DARPA 1000-words English language Resource Managemen (RM)]{RM}{RM}{DARPA 1000-words English language Resource Managemen}
\newacronym[plural=Aurora4,firstplural=Aurora 4 Task (Aurora4)]{Aurora4}{Aurora4}{Aurora 4 Task}
\newacronym[plural=WSJ,firstplural=Wall Street Journal (WSJ)]{WSJ}{WSJ}{Wall Street Journal}

\DeclareFloatVCode{somespace}{\vspace{1.667\baselineskip}}
\title{Investigation of Densely Connected Convolutional Networks with Domain Adversarial Learning for Noise Robust Speech Recognition}

\author{Chia Yu Li, Ngoc Thang Vu}
\affil{Institute for Natural Language Processing, University of Stuttgart, Germany}
\email{\{licu,thangvu\}@ims.uni-stuttgart.de}

\begin{document}
\selectlanguage{english}

\maketitle

\begin{abstract}
We investigate \glspl{DenseNets} and their extension with domain adversarial training for noise robust speech recognition. \glspl{DenseNets} are very deep, compact convolutional neural networks which have demonstrated incredible improvements over the state-of-the-art results in computer vision. 
Our experimental results reveal that \glspl{DenseNets} are more robust against noise than other neural network based models such as deep feed forward neural networks and convolutional neural networks. Moreover, domain adversarial learning can further improve the robustness of DenseNets against both, known and unknown noise conditions.
\end{abstract}

\section{Introduction}
In the last years, automatic speech recognition (ASR) performance has been significantly improved through the use of neural networks with deep structures~\cite{deep-neural-networks-for-acoustic-modeling-in-speech,conversational-speech-transcription-using-context-dependent-deep-neural-networks-2,context-dependent-pre-trained-deep-neural-networks-for-large-vocabulary-speech-recognition}. Various neural network architectures have been developed to improve ASR performance. They are variations of \glspl{TDNNs}~\cite{waibel1990phoneme}, \glspl{CNNs} ~\cite{abdel2012applying}  \glspl{RNNs}~\cite{graves2013speech}, and their combinations ~\cite{convolutional-networks-for-images-sppech-and-time-series}. 
Among them, very deep \glspl{CNNs}  demonstrate impressive performance  ~\cite{very-deep-convolutional-neural-networks-for-robust-speech-recognition,deep-convolutional-neural-networks-layer-wise-context-expansion-attention,xiong2018microsoft} especially in noisy conditions ~\cite{very-deep-convolutional-neural-networks-for-robust-speech-recognition}. 

Recently in the computer vision research community, densely connected convolutional networks (DenseNets) have obtained significant improvements over the state-of-the-art networks on four highly competitive object recognition benchmark tasks~\cite{DenseNet}. The idea is to introduce shorter connections between layers close to the input and those close to the output which alleviate the vanishing-gradient problem. Furthermore, \glspl{DenseNets} require fewer parameters than traditional \glspl{CNNs} with the same deep structure~\cite{DenseNet}. In \cite{DenseNetASR}, we showed that \glspl{DenseNets} can be used for acoustic modeling achieving impressive performance.

In this paper, we explore noise robustness of \glspl{DenseNets} and their extension with domain adversarial learning. This method was originally proposed by Ganin et al. \cite{ganin2016domain} for unsupervised domain adaptation in natural language processing and was then applied to deep feed forward neural networks (DNNs) for noise robust speech recognition \cite{Adversarial-Multi-task-Learning-of-Deep-Neural-Networks-for-Robust-Speech-Recognition, DomainAD}. However to the best of our knowledge, domain adversarial learning has never been examined with a complex network like \glspl{DenseNets} before. 
Our experimental results on noisy data demonstrate that \glspl{DenseNets} can effectively improve the noise robustness of the system outperforming other neural based models. Using domain adversarial learning can further improve their robustness against both, known and unknown noise conditions. 

\section{Methods}
\subsection{\glspl{DenseNets} Acoustic Models}
In this subsection, we first describe \glspl{DenseNets} and review the method how to use \glspl{DenseNets} for acoustic modeling \cite{DenseNetASR}. The key idea of \glspl{DenseNets} is the introduction of shorter connections between layers close to the input and those close to the output which alleviate the vanishing-gradient problem. 
For acoustic modeling, \glspl{DenseNets} take the unadapted features 40-dimensional log Mel filterbank as input and predict the context dependent HMM states (senones) \cite{DenseNetASR}.

\begin{figure}[!hbt]
  \centerline{\includegraphics[width=.5\columnwidth]{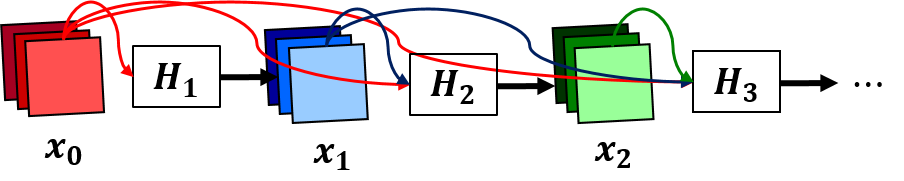}}
  \caption{A 3-layer \text{\it dense block}, in which each layer takes all preceding feature-maps as input}
  \label{fig:Denseblock}
\end{figure}

Given an input $x_0$ and a CNN with $N$ layers, where each layer $n$ is equipped with a nonlinear transformation $H_n(\cdot)$ which is the composition of three consecutive operations: batch normalization, followed by a ReLU and a $3 \times 3$ convolution, \glspl{DenseNets} introduce direct connections from any layer to all subsequent layers. The output of $n^{th}$ layer is:
\begin{equation}
x_n = H_n([x_0,x_1,x_2,...,x_{n-1}])
\label{Hn}
\end{equation}
where $[x_0,x_1,x_2,...,x_{n-1}]$ refers to the concatenation of the feature maps yielded in all the previous layers. Fig.~\ref{fig:Denseblock} illustrates the dense connectivity structure, in which each layer takes all preceding feature-maps as input. This structure is called \text{\it dense block}.

\begin{figure*}[!hbt]
  \centering
  \includegraphics[width=0.9\columnwidth]{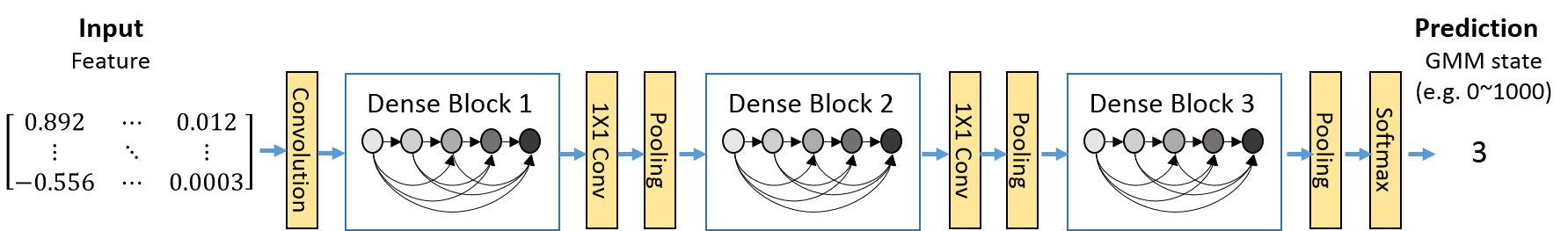}
  \caption{A DenseNet architecture with three dense blocks connected via transition layers}
  \label{fig:DenseNet3b}
\end{figure*}

\glspl{DenseNets} consist of multiple \text{\it dense} \text{\it blocks}, connected in series and separated by \textit{transition layers}.
Each transition layer consists of a $1 \times 1$ convolution layer and a $2 \times 2$ average pooling layer.
Fig.~\ref{fig:DenseNet3b} illustrates how these \text{\it dense blocks} and \text{\it transition layers} are composed in \glspl{DenseNets}. Note that pooling is only performed outside of \text{\it dense blocks}.

Furthermore, \glspl{DenseNets} reduce the number of feature-maps by $1 \times 1$ convolution layer in the \text{\it transition layer} to improve model compactness. For example, if a \text{\it dense block} has $y$ feature-maps, the transition layer generates $\lfloor \theta y \rfloor$ output feature-maps, where $\theta$ is the compression factor and the range is $0 <$ $\theta$ $\leq 1$.
The \textit{growth rate} of \glspl{DenseNets} is the number of channels in their convolution layers. By equation~\eqref{Hn}, the $n^{th}$ layer within a dense block  has $k \times (n-1) + k_0$ input feature-maps, where $k_0$ is the number of input channels and $k$ (the model's growth rate) is the number of channels for subsequent convolution layers. \glspl{DenseNets} have better performance when $k$ is a small integer, e.g.\ $k = 12$ \cite{DenseNet}.




\subsection{Domain Adversarial Learning}
In this subsection, we introduce the extension of \glspl{DenseNets} with domain adversarial learning \cite{ganin2016domain}. In this context, noisy conditions act as domain information.

\begin{figure}[t]
  \centerline{\includegraphics[width=0.4\columnwidth]{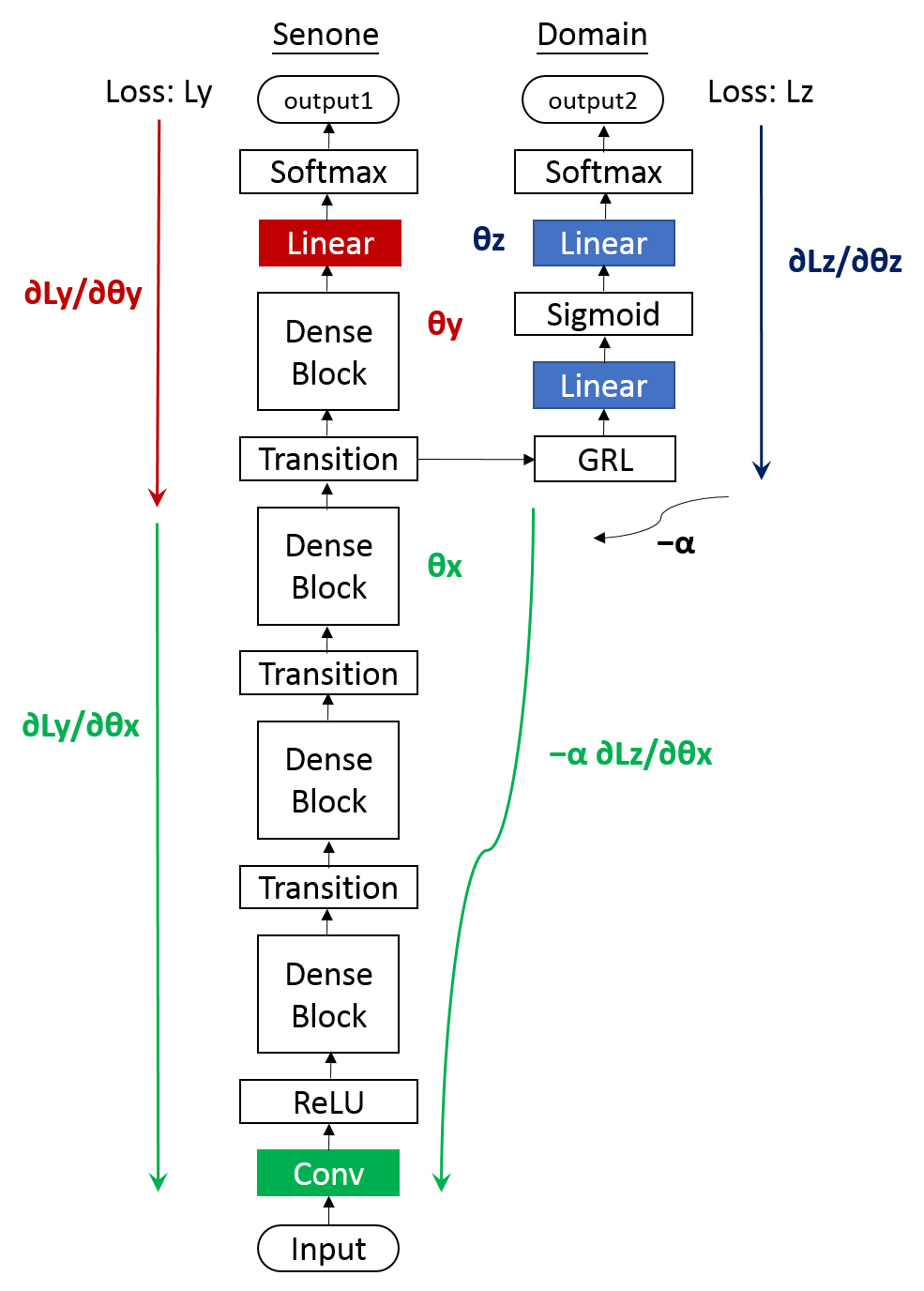}}
  \caption{An architecture of DenseNets with domain adversarial learning which consists of there sub-networks: features extractor sub-network (x), senone classification sub-network (y) and domain classification sub-network (z)}
  \label{fig:ad_densenet}
\end{figure}

The overall architecture of the domain adversarial learning of \glspl{DenseNets} is shown in Fig~\ref{fig:ad_densenet}. It consists three sub-networks: the sub-network ($y$) is for senone classification, the sub-network ($z$) is for domain classification and the share-network ($x$) is the share part of the two tasks. The shared-network ($x$) can be seen as a feature extractor to convert an input vector to its latent representation. Each output sub-network acts as a classifier to calculate posterior probabilities of classes given the this latent representation \cite{Adversarial-Multi-task-Learning-of-Deep-Neural-Networks-for-Robust-Speech-Recognition, DomainAD}. In the domain adversarial learning, the representation is learned adversarially to the domain classification and friendly to the senone classification, so that domain-dependent information to the senone classifier is removed from the representation.

Let $\theta_x, \theta_y, \theta_z$ denote the parameters of the share-network (x), sub-network (y) and sub-network (z), respectively. The cross-entropy loss functions for the senone classifier and domain classifier are defined as 

\begin{equation}
\mathcal{L}_y(\theta_x, \theta_y) = - \sum_{i} logP(y_i|x_i;\theta_x,\theta_y)
\label{ADY}
\end{equation}
\begin{equation}
\mathcal{L}_z(\theta_x, \theta_z) = - \sum_{i} logP(z_i|x_i;\theta_x,\theta_z)
\label{ADZ}
\end{equation}

\noindent The parameters are updated as
\begin{equation}
\theta_y \leftarrow \theta_y - \epsilon \dfrac{\partial\mathcal{L}_y}{\partial\theta_y}
\label{updatey}
\end{equation}
\begin{equation}
\theta_z\leftarrow \theta_z - \epsilon \dfrac{\partial\mathcal{L}_z}{\partial\theta_z}
\label{updatey}
\end{equation}
\begin{equation}
\theta_x\leftarrow \theta_x - \epsilon (\dfrac{\partial\mathcal{L}_y}{\partial\theta_x} - \lambda \dfrac{\partial\mathcal{L}_z}{\partial\theta_x})
\label{updatex}
\end{equation}

\noindent where $\lambda$ is the gradient reversal coefficient which is a positive scalar parameter to adjust the strength of the regularization. 

The first layer of the sub-network ($z$) is the gradient reversal layer (GRL) as proposed in \cite{Adversarial-Multi-task-Learning-of-Deep-Neural-Networks-for-Robust-Speech-Recognition}. In the forward propagation phase, the GRL just passes the input to the output as follows:
\begin{equation}
\xi_{out}\leftarrow \xi_{in}
\label{rgl}
\end{equation}
where $\xi_{in}$ and $\xi_{out}$ represent input and output vectors of the layer, respectively.
In the backward propagation phase, the GRL reverses the gradient by multipling it with $\lambda$ as follows:
\begin{equation}
\dfrac{\partial\mathcal{L}}{\partial\xi^{in}}\leftarrow -\lambda\dfrac{\partial\mathcal{L}}{\partial\xi^{out}}
\label{rgl-2}
\end{equation}

Hence the shared-network (x) is trained adversarially to the sub-network (y) for the domain classification. When the training is finished, the output of the entire network (x, y) for the senone classification is used for decoding.

\section{Setup}
Two experiments are conducted in this paper. The goal for the first experiment is to explore \glspl{DenseNets}' robustness at different levels of noise. We compare the baseline models, which are \glspl{DNNs}, \glspl{CNNs} and \glspl{TDNNs}, and \glspl{DenseNets} on noise corrupted \glspl{RM}. The second experiment examines the effectiveness of domain adversarial learning, in which we compare the performance of \glspl{DenseNets}, DenseNets-AD and the best baseline model \glspl{TDNNs} on noise corrupted \glspl{RM} and \glspl{Aurora4}. 

\subsection{Resources}
\subsubsection{Data}
\begin{figure}[!htb]
\minipage{0.32\textwidth}
  \includegraphics[width=0.8\linewidth]{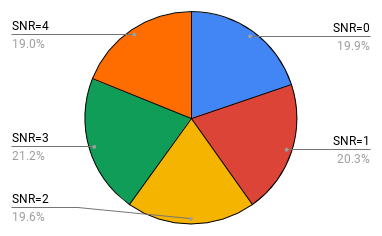}
\endminipage\hfill
\minipage{0.32\textwidth}
  \includegraphics[width=0.8\linewidth]{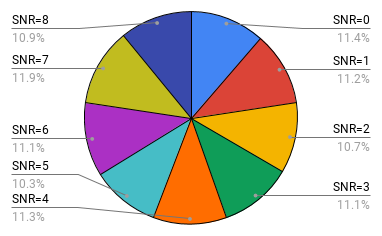}
\endminipage\hfill
\minipage{0.32\textwidth}%
  \includegraphics[width=0.8\linewidth]{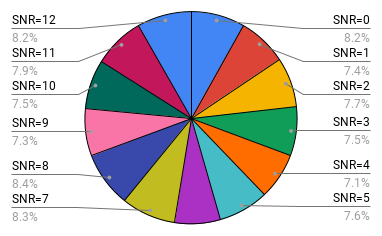}
\endminipage
\caption{The data composition of Data-1, Data-2 and Data-3 from left to right.}\label{fig:data}
\end{figure}

The noise corrupted \glspl{RM} is made by artificially adding different types of noise at different values of SNR \footnote{Signal-to-Noise ratio (SNR) is defined as the ratio of the power of a signal to the power of noise ($\frac{P_{signal}}{P_{noise}}$)} to \glspl{RM} \cite{RM} using the large-scale open-source acoustic simulator developed in \cite{C2}. It contains 1,993 noisy conditions: 1,500 are used for training and 493 for testing. 
We created three noise corrupted data sets with different SNRs: Data-1 (SNR from 0 to 4), Data-2 (SNR from 0 to 8) and Data-3 (SNR from 0 to 12). Figure~\ref{fig:data} shows the data distribution of Data-1, Data-2 and Data-3. For example, 19.9\% utterances in Data-1 are adding noise (randomly chosen) at SNR=0, 20.3\% of them are at SNR=1, 19.6\% of them are at SNR=2, 21.2\% of them are at SNR=3 and 19\% of them are at SNR=4. 
Two noise corrupted test sets are used in this paper. 
In the "known-noise test set" (KNN), the noise is randomly picked from 1,500 training noise and added to the utterance at the same range of SNR used in the training set. On the contrary in the "unknown-noise test set" (UKN), the noise is selected from 493 testing noises and added to the utterance.

The Aurora 4 task, which is a medium vocabulary task speech recognition task, is based on the Wall Street  Journal (WSJ0) dataset \cite{AURORA4}. It contains 16 kHz speech data in the presence of six additive  noises (car, crowd of people, restaurant, street, airport and train station) with linear convolutional channel distortions. The multi-condition training set with 7138 utterances from 83 speakers includes a combination of clean utterance and utterance corrupted by one of six different noises at 10-20 dB SNR. 
3,569 utterances are from the primary Sennheiser micro-phone and 3,569 utterances are from the secondary microphone.  The test data is made using the same types of noise and microphones, and these can be classified into five test-conditions: clean, noisy, clean  with  channel distortion, noisy with channel distortion, and all of them which will be referred to as A, B, C, D and Average respectively.

\subsubsection{Baseline systems}
The baseline models in our experiments are \glspl{DNNs}, \glspl{CNNs} and \glspl{TDNNs}. \glspl{DNNs} take the 40-dimensional log Mel filterbank features as input and contain six hidden layers with sigmoid activation functions and one fully-connected output layer with a softmax activation. Each hidden layer has 1024 units. \glspl{CNNs} are composed of two convolution layers and max-pooling layers, and four affine layers with sigmoid activation function. Each affine layer contains 1024 units and is trained using 40-dimensional log Mel filterbank features with the first and the second time derivatives. Both \glspl{DNNs} and \glspl{CNNs} use the same context window of five and batch size of 256. The best \glspl{TDNNs} in Kaldi use in addtion iVector for speaker adapted systems. They contain five weight layers with different context specifications (subsampling). Furthermore, the \glspl{TDNNs} recipe applies data augmentation technique which does speed perturbation of the training data in order to emulate vocal tract length perturbations and speaking rate perturbation. All the ASR systems are built up with the Kaldi speech recognition toolkit ~\cite{kaldi}. The acoustic models except TDNNs are implemented with PDNN (A Python Deep learning toolkit) ~\cite{kaldi-pdnn}, Theano ~\cite{theano}, Lasagne ~\cite{lasagne} and \glspl{DenseNets} source code ~\cite{DenseNet}.

\subsubsection{Hyperparameters for \glspl{DenseNets} and DenseNets-AD}
The architecture of \glspl{DenseNets} in this paper is the same as the best model in the previous work ~\cite{DenseNetASR}: the first layer is $3 \times 3$ convolution which is followed by 4 dense blocks. Each block contains 14 $3 \times 3$ convolutional layers. Each dense block except the last one is followed by a transition which consists of $1 \times 1$ convolution and $2 \times 2$ average pooling. The depth is 65, the growth rate is 12 and the compression ratio is 0.5. All the convolutional layer use kernel size $3X3$. The architecture of DenseNets-AD is mentioned in figure~\ref{fig:ad_densenet} except the shared layer is only the first convolutional layer and the following layers are training for senone (phoneme) recognition task. The gradient reversal coefficient $\lambda$ is 0.5. The input features for both models is 40-dimensional log Mel filterbank features with the first and the second time derivatives.

\begin{figure}[!htb]
  \centering
  \includegraphics[scale=.4]{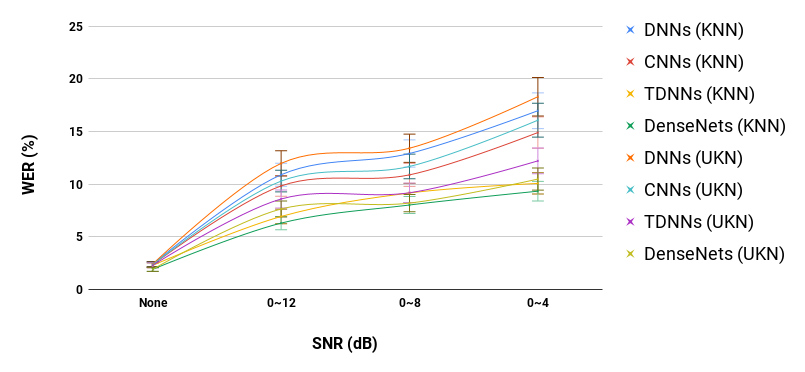}
  \caption{The WERs of TDNN, DenseNets and DenseNets-AD on noisy test sets at different SNR ranges; DNNs(KNN) means the WER of DNNs on known-noise test set (KNN) and DNNs(UKN) means the WER of DNNs on unknown-noise test set (UKN).}
  \label{fig:trend}
\end{figure}

\section{Results and Discussion}
Figure~\ref{fig:trend} shows the WERs of all the baseline models and \glspl{DenseNets} on noise corrupted \glspl{RM} test set at different ranges of SNR. Note that all the models are examined on the noise corrupted test set at the same range of SNR as the training. For example, the model are trained on the noise corrupted training set at the SNR range from 0 to 4 and tested on the noise corrupted test set at the same SNR range. The experimental results in figure~\ref{fig:trend} show that the WERs of \glspl{DNNs} and \glspl{CNNs} increase when the SNR decreases. However, \glspl{TDNNs} and \glspl{DenseNets} are relatively stable when reducing SNR. Overall, \glspl{DenseNets} outperform the baseline models on all the test sets. 

\begin{table}[t]
\small
    \begin{tabular}{|l|l|c|c|}\hline
    \textbf{SNR} & \textbf{System} & \textbf{KNN}& \textbf{UKN} \\ \hline
    &TDNNs	  &6.93 &8.59\\ 
     0-12&DenseNets	  &6.30 &7.64\\ 
    &DenseNets-AD	  &6.11&6.97\\
    \hline
     &TDNNs	  &9.16 &9.18\\ 
    0-8&DenseNets-AD  & 8.02&8.20\\
    &DenseNets-AD	  & 7.84&7.95\\
    \hline
    &TDNNs	  &10.07 &12.21\\
    0-4&DenseNets	  &9.33 &10.48\\ 
    &DenseNets-AD	  &8.64&9.68\\
    \hline
	\end{tabular}
    \caption{\label{RM}The WERs of TDNN, DenseNets and DenseNets-AD on KNN and UKN test sets. Where KNN means known-noise test set and UKN means uknow-noise test set.}
\end{table}

\begin{table}[t]
\small
  \begin{tabular}{|l|c|c|c|c|c|c|}
    \hline
    \textbf{System} & \textbf{A} & \textbf{B}& \textbf{C}& \textbf{D} & Average\\
    \hline
    TDNNs & 3.47  &7.44 &10.14 &21.91 & 13.57 \\
    DenseNets & 3.57  & 7.29 & 7.12& 16.56 & 11.53\\
    DenseNets-AD & 3.58  & 6.58 & 6.76 & 16.42 & 10.21\\
    \hline
  \end{tabular}
  \caption{\label{tabble:AURORA4}WERs of TDNNs, DenseNets and DenseNets-AD on Aurora4 test set with A, B, C, D conditions. (A: clean and Sennheiser mic, B: Sennheiser mic and noise added, C: clean and 2nd mic, D: 2nd mic and noise added)}
\end{table}

Table~\ref{RM} and Table~\ref{tabble:AURORA4} show the comparison between \glspl{TDNNs}, \glspl{DenseNets} and \glspl{DenseNets-Dal} on the noise corrupted \glspl{RM} and \glspl{Aurora4}. As expected, the WERs of three models increase when the SNR decreases. However,  \glspl{DenseNets-Dal} achieves best performance on both KNN and UKN test sets at all three SNR ranges. One of the reason is that TDNNs and \glspl{DenseNets} recognize noise as speech when the noise becomes severe. Table~\ref{ex1} shows one example of TDNNs and \glspl{DenseNets} failing to distinguish the noise and speech while DenseNets-AD was unaffected.  

\floatplacement{figure}{H}

\begin{table}[!htb]
\small
  \begin{tabular}{|l|l|}\hline
    Reference& LIST FULL LOCATION DATA FOR TRACK FFF088 \\
    TDNNs& LIST FULL LOCATION DATA FOR TRACK FFF088 \text{\color{red}TO EIGHT}\\
    DenseNets &LIST FULL LOCATION DATA FOR TRACK FFF088 \text{\color{red}IN THE EIGHT}\\
    DenseNets-AD &LIST FULL LOCATION DATA FOR TRACK FFF088\\
    \hline
  \end{tabular}
  \caption{\label{ex1}The references and the ASR outputs from TDNNs, DenseNets and \glspl{DenseNets-Dal} of an utterance when adding dumpster truck noise at SNR=1}
\end{table}




\section{Conclusions}
This paper investigates noise robustness of \glspl{DenseNets} and their extension with domain adversarial learning. Our experimental results demonstrate that \glspl{DenseNets} are more robust against noise than other types of neural networks. Furthermore, we show that applying domain adversarial learning improves the performance of \glspl{DenseNets} and model generalization.

\newpage
\bibliographystyle{essv}
\bibliography{essv}

\end{document}